\documentclass[pdflatex,sn-mathphys-num,oneside]{sn-jnl}

\usepackage{graphicx}
\usepackage{amsmath}
\usepackage{amssymb}
\usepackage{amsfonts}
\usepackage{mathrsfs}
\usepackage[title]{appendix}
\usepackage{xcolor}
\usepackage{textcomp}
\usepackage{manyfoot}
\usepackage{booktabs}
\usepackage{multirow}
\usepackage{array}
\usepackage{url}
\usepackage{algorithm}
\usepackage{algorithmic}

\makeatletter
\def\p@decl{}
\def\p@ethics{}
\def\p@funding{}
\makeatother

\begin{document}

\title[SAP-Depth]{SAP-Depth: Scene-Adaptive Probability Framework for Self-Supervised Monocular Depth Estimation}

\author[1]{\fnm{Bowen} \sur{He}}\email{15155651510@163.com}
\author*[1]{\fnm{Li} \sur{Zhang}}\email{lizhang@hfut.edu.cn}
\author*[1]{\fnm{Yujie} \sur{Chen}}\email{2023111347@mail.hfut.edu.cn}
\author[1]{\fnm{Chang} \sur{Cheng}}\email{chengchang58@126.com}

\affil[1]{\orgname{Hefei University of Technology}, \orgaddress{\city{Hefei}, \country{China}}}

\abstract{Self-supervised monocular depth estimation remains challenging due to the loss of fine-grained details and limited adaptability to varying scene structures. In this paper, we propose SAP-Depth, a novel architecture based on Scene-Adaptive Probabilistic. Unlike traditional methods, we introduce a Dynamic Bins mechanism powered by a Learnable Parameter Predictor, which adaptively estimates scene-specific depth distributions conditioned on the input image context. Furthermore, a Bin-wise Probability Distribution Decoder is designed to effectively fuse multi-scale representations to predict dense depth probability distributions, significantly enhancing boundary sharpness and structural recovery. Through the collaborative coupling of these proposed modules, our model achieves dynamic prediction of scene-specific depth distributions. Extensive experiments on the KITTI dataset show that SAP-Depth outperforms existing state-of-the-art methods in accuracy, particularly in its ability to better preserve complex edge details. Further, tests on the Make3D dataset validate the model's excellent generalization capability and robustness.}


\maketitle

\section{Introduction}

Monocular depth estimation is a fundamental and critical task in computer vision, aiming to recover 3D scene depth information from a single RGB image. This technology is core to numerous applications, including autonomous driving, 3D reconstruction~\cite{ref2}, Augmented Reality (AR)~\cite{ref3}, and robotic navigation~\cite{ref4}. However, inferring 3D depth from 2D imagery is an inherently ill-posed problem, plagued by scale ambiguity and geometric challenges.

Although the advent of deep learning has significantly advanced monocular depth estimation, the early supervised learning paradigm heavily relies on dense ground-truth depth data~\cite{ref9,ref10,ref22}. To alleviate the dependence on annotated data, self-supervised learning has emerged as an attractive alternative. These methods formulate depth estimation as a novel view synthesis problem, training by minimizing photometric reconstruction loss without requiring dense ground truth~\cite{ref5, ref11}. Pioneering works~\cite{ref5, ref11} established a robust framework that jointly trains depth and pose networks, leveraging geometric constraints from adjacent video frames or stereo pairs and adopting photometric reprojection error as the sole supervision signal. Mainstream self-supervised methods typically treat the task as a pixel-wise continuous regression problem, which often leads to over-smoothed depth predictions and fails to preserve object boundaries~\cite{ref11, ref14}. Designing a model that concurrently achieves the data efficiency of self-supervision and the high accuracy of supervised paradigms remains an open challenge. 

\begin{figure*}[!t]
    \centering
    \includegraphics[width=0.92\linewidth]{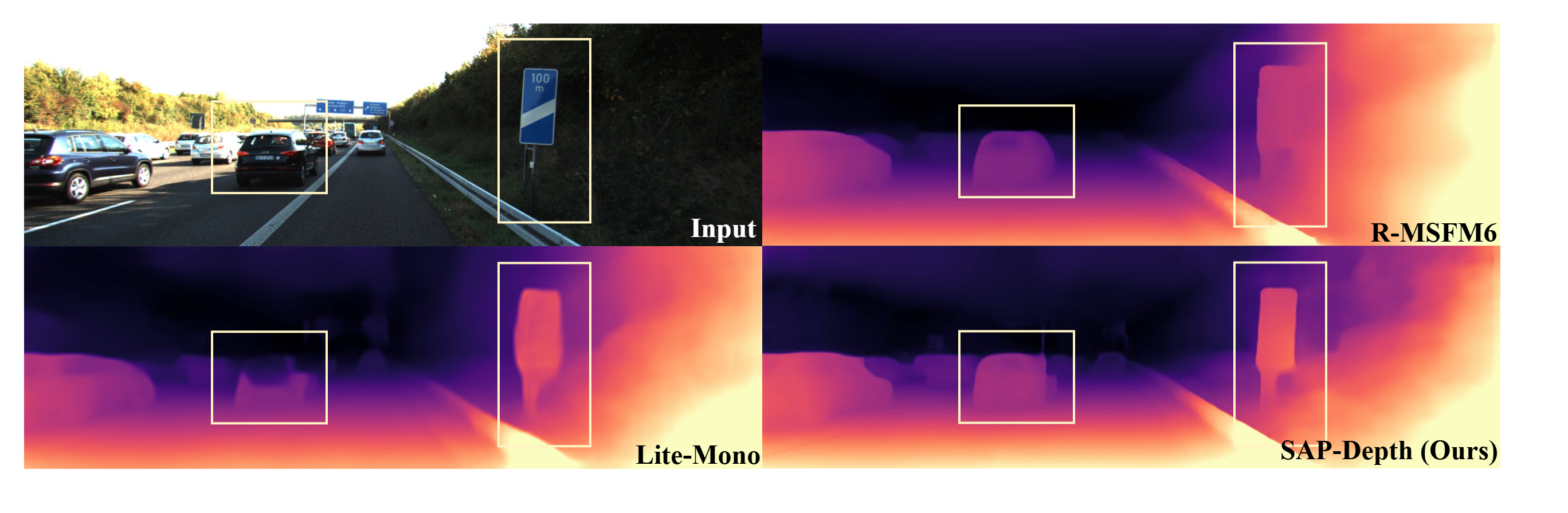}
    \caption{The proposed SAP-Depth generates more accurate depth maps with sharper boundaries compared to Lite-Mono\cite{ref12}, R-MSFM6\cite{ref34}. Ours shows better contour details.} 
    \label{fig:teaser}
    \vspace{-0.4cm}
\end{figure*}

To address these limitations, we propose a scene-adaptive probability framework based on dynamic depth bins, aiming to enrich existing frameworks. Unlike traditional regression methods, we formulate depth estimation as a "classification-regression" task, enabling effective end-to-end training without depth ground truth to generate more accurate depth maps. To enable self-supervised monocular depth estimation to learn from more accurate reprojected images as supervisory signals, we meticulously design a Bin-wise Probability Distribution (BPD) decoder. Combined with depth bins, this design allows our framework to adaptively generate appropriate depth intervals and their corresponding probability distributions for diverse scenes. 

The main contributions of this paper are as follows:
\begin{itemize}
    \item We propose SAP-Depth, a multi-scale perception framework that leverages low-resolution deep semantic features to guide depth binning prediction. This approach effectively addresses boundary blurring and structural over-smoothing, significantly sharpening object contours. 
    \item We design a Bin-wise Probability Distribution (BPD) decoder, which integrates multi-scale feature fusion with an adaptive probabilistic bin strategy. This design effectively mitigates the loss of fine-grained details inherent in deterministic regression paradigms.  
    \item Extensive experiments on the KITTI dataset demonstrate that the proposed method achieves state-of-the-art performance without any auxiliary data or depth ground truth. Furthermore, the model exhibits excellent zero-shot generalization capabilities on the Make3D dataset, fully validating its robustness and generality across scenes.
\end{itemize}

\section{Related Work}

\subsection{Supervised Monocular Depth Estimation}
Supervised monocular depth estimation typically uses an encoder-decoder architecture to directly regress depth values~\cite{ref9,ref10,ref22}. While recent advancements in Vision Transformers for dense prediction (ViT)~\cite{ref16}, diffusion models (e.g., Marigold~\cite{ref18}), and large-scale data-driven models (e.g., Depth Anything~\cite{ref45}) have achieved state-of-the-art results by enhancing global context modeling and scaling up training distributions, they still fundamentally rely on extensive depth ground-truth priors and exhibit relatively high computational complexity.

\subsection{Self-Supervised Monocular Depth Estimation}
To alleviate the need for dense annotations, self-supervised methods formulate depth estimation as a novel view synthesis problem, optimizing photometric reprojection loss from stereo pairs or monocular videos~\cite{ref13,ref17,ref33}. Recent studies have further improved performance by designing detail-enhancing modules~\cite{ref30} or incorporating diffusion models to refine boundaries~\cite{ref21}. However, without explicit ground-truth constraints, mainstream self-supervised models that treat the task as continuous regression still struggle with over-smoothed outputs and blurred object boundaries.

\subsection{Adaptive Bins and Probabilistic Regression}
In supervised learning, the strategy of reformulating depth prediction as a "classification-regression" task using adaptive binning has been shown to estimate high-quality dense depth maps from a single RGB input image~\cite{ref8, ref9, ref10, ref22, ref43}. However, transferring this paradigm to a self-supervised setting is highly challenging, primarily due to the inability to use ground truth to guide the bin division. Our framework addresses this by explicitly integrating a Bin-wise Probability Distribution (BPD) decoder based on a multi-scale backbone. This design enables effective end-to-end self-supervised training, significantly alleviating core issues like over-smoothing and boundary loss without relying on implicit cost volumes or enhanced pose networks.

\section{Proposed Method}
\label{sec:method}

In this section, we elaborate on the details of our proposed self-supervised monocular depth estimation framework. Our proposed Bin-wise Probability Distribution (BPD) decoder integrates multi-scale feature fusion with an adaptive probabilistic binning strategy to adaptively predict depth information across diverse scenes. This effectively addresses the issues of fine-grained detail degradation and the insufficient adaptability of models to varied scene structures. We first present the overall pipeline of the system in Section \ref{sec3.1}. Subsequently, we detail the architecture of the Depth Network in Section \ref{sec3.2}. Finally, we describe the Pose Network and the self-supervised training in Section \ref{sec3.3}.

\subsection{Design Motivation}\label{sec3.1}

Existing self-supervised monocular depth estimation methods are plagued by two key limitations: they often lose fine-grained structural details (e.g., object edges and texture boundaries) during depth regression, and they lack adaptive modeling capabilities to handle the diversity of real-world scenes with varying layouts, depth ranges, and object scales. To address these issues, we propose SAP-Depth, which integrates a Hierarchical Adaptive Probabilistic Feature Aggregation Module (HAP-FAM) with a core Learnable Parameter Predictor. This predictor dynamically captures scene-specific semantic and geometric features from input images, generating adaptive depth modeling parameters tailored to each scene, and thus enables the network to predict denser, more accurate depth maps with well-preserved structural details. 

\begin{figure*}[!t]
    \centering
    \includegraphics[width=0.92\linewidth]{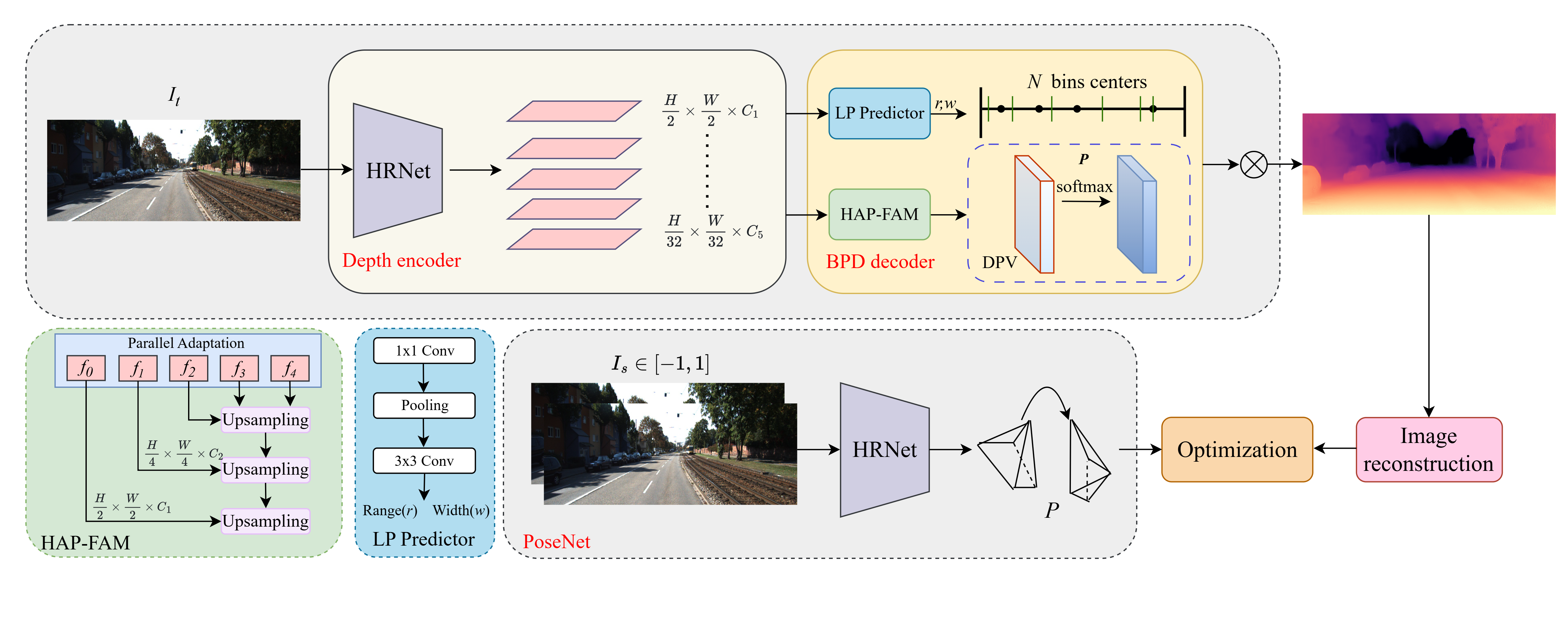} 
    \caption{\textbf{Overview of our proposed SAP-Depth framework.} The system consists of two main components: a depth network and a pose network. (a) The depth network utilizes a shared encoder to extract multi-scale features, which are then decoupled into two branches: the LP Predictor predicts global adaptive depth bins, while the HAP-FAM estimates pixel-wise bin probabilities. The final high-quality depth map is generated by the linear combination of the predicted bins and probabilities. (b) During training, the pose network takes adjacent frames ($I_t, I_{t'}$) to estimate the 6-DoF relative pose $T_{t \to t'}$, enabling the photometric reconstruction loss for self-supervision.}
    \label{fig:overview}
    \vspace{-0.4cm}
\end{figure*}

\subsection{Depth Network Architecture}\label{sec3.2}

{\textbf{Depth encoder.}} 
We adopt HRNet\cite{ref7} as the depth encoder, which takes a single RGB image $I_t$ as input and extracts a set of multi-scale features $F = \{f_i\}_{i=0}^{4}$.

{\textbf{Depth decoder.}} 
We design a Bin-wise Probability Distribution (BPD) decoder through the integration of multi-scale feature fusion and an adaptive probabilistic binning strategy. This architecture is composed of three synergistic components, namely the Learnable Parameter Predictor (LP Predictor), the Hierarchical Adaptive Probabilistic Feature Aggregation Module (HAP-FAM), and the Depth Probabilistic Volume (DPV). The decoder takes multi-scale features from the HRNet encoder as input to dynamically aggregate global and local contextual information. Finally, it outputs dense depth maps via a linear combination of probability distributions and the centers of depth bins, which effectively recovers continuous depth values from the discrete probability volumes. 

\textbf{Theoretical Advantage of Probabilistic Bins.} Mathematically, formulating depth estimation as a classification-regression task via dynamic bins presents profound optimization advantages over direct continuous regression. Direct regression relies on minimizing photometric error over depth $D$, a landscape that is highly non-convex due to local texture ambiguities and occlusions. By evaluating probabilities across discrete depth bins and aggregating them via expectation, the BPD decoder essentially performs a continuous relaxation of the non-convex objective. This mechanism softens local minima, leading to a much smoother optimization trajectory. Furthermore, standard continuous regression inherently assumes a unimodal distribution of depth uncertainty, causing severe "flying pixel" artifacts around object boundaries where depth jumps abruptly. Conversely, our probability volume explicitly models multimodal uncertainty, assigning sharp probabilities at discontinuous edges without averaging foreground and background. This leads to the remarkable boundary-preserving capability of our method. The detailed forward process is formalized in Algorithm \ref{alg:sap_depth}.

\begin{algorithm}[htbp]
\caption{Self-Supervised Monocular Depth Estimation via Dynamic Bins}
\label{alg:sap_depth}
\begin{algorithmic}[1]
\REQUIRE Unlabeled monocular image sequence $\mathcal{S} = \{I_{t-1}, I_t, I_{t+1}\}$, Intrinsic matrix $K$, Number of bins $N$.
\STATE Initialize HRNet Encoder $\mathcal{E}$, LP Predictor $\mathcal{P}_{bin}$, BPD Decoder $\mathcal{D}$, and Pose Net $\mathcal{P}_{pose}$ with parameters $\theta$.
\FOR{each training batch in dataset}
    \STATE \verb|// Phase 1: Multi-scale Feature Extraction|
    \STATE Extract multi-scale dense features: $F_{t} = \mathcal{E}(I_t)$
    
    \STATE \verb|// Phase 2: Dynamic Bins Generation (Macro)|
    \STATE Predict global range bounds and normalized widths: 
    \STATE \hspace{0.5cm} $(d_{min}, d_{max}, \mathbf{w}) = \mathcal{P}_{bin}(F_{t})$
    \STATE Compute adaptive bin centers $\mathbf{c} = \{c_1, c_2, \dots, c_N\}$ via Eq. (1) and (2).
    
    \STATE \verb|// Phase 3: Dense Probability and Depth (Micro)|
    \STATE Generate multi-scale probability volumes: 
    \STATE \hspace{0.5cm} $\mathbf{P} = \mathcal{D}(F_{t}, \mathbf{c})$ \quad \textit{(where $\sum_{i=1}^N P_i^{(u,v)} = 1$)}
    \STATE Compute mathematical expectation for continuous depth $D_t$:
    \STATE \hspace{0.5cm} $D_t^{(u,v)} = \sum_{i=1}^{N} P_i^{(u,v)} \cdot c_i$
    
    \STATE \verb|// Phase 4: Self-Supervised Optimization|
    \STATE Estimate relative 6-DoF poses: $T_{t \to t'} = \mathcal{P}_{pose}(I_t, I_{t'})$
    \STATE Synthesize target image $\hat{I}_{t'}$ using $D_t$, $T_{t \to t'}$, $K$, and $I_{t'}$
    \STATE Compute photometric loss and edge-aware smoothness loss
    \STATE Update network parameters $\theta$ via backpropagation
\ENDFOR
\end{algorithmic}
\end{algorithm}

{\textbf{LP Predictor.}} 
The LP Predictor fuses multi-scale features to estimate dynamic depth ranges and bin distributions. It first standardizes input channels via 1$\times$1 convolutions, upsamples and concatenates them to the finest resolution, and then compresses the fused features into a global vector via adaptive pooling. Finally, two parallel heads generate normalized bin widths and Sigmoid-activated depth range parameters, respectively. Specifically, to filter out redundant low-level details and focus on scene semantics, it extracts the top three high-level multi-scale features from the HRNet encoder (with channel dimensions of 36, 72, and 144). The initial 1$\times$1 convolutions standardize these features to a uniform 64 channels, which are then concatenated into a 192-channel fused tensor before being compressed back to 64 channels and pooled.

{\textbf{HAP-FAM}.}
First, Hierarchical Adaptive Probabilistic Feature Aggregation Module (HAP-FAM) processes encoder-derived multi-scale features with parallel convolutional blocks and 1$\times$1 convolutional layers to unify channel dimensions, then upsamples/downsamples features to align resolutions, and aggregates features across scales through additive fusion at multiple hierarchical levels. Specifically, it operates on all five multi-scale features (with original channels of 64, 18, 36, 72, and 144). By executing progressive additive fusion across four hierarchical stages, the module effectively embeds high-frequency structural details from shallow layers (e.g., 64 and 18 channels) into the semantic representations of deeper layers, ultimately projecting them to a 16-channel bottleneck representation before entering the probability decoder. Additionally, the module outputs two sets of scene-level parameters (shared across the entire image): normalized depth bin widths and depth range parameters (min/max depth), which serve as the basis for subsequent depth bin center calculation:
\begin{enumerate}
    \item Normalized Range Parameters ($r \in \mathbb{R}^2$): Values constrained within $[0, 1]$, used to determine the $d_{\min}$ and $d_{\max}$ of the current scene.
    \item Normalized Widths ($w \in \mathbb{R}^N$): A vector defining the relative width distribution of $N$ depth bins, where $N$ is the preset number of bins.
\end{enumerate}

{\textbf{DPV.}} 
The Depth Probabilistic Volume (DPV) leverages the enriched fused features to accurately compute depth bin probabilities (via softmax over logits generated by a 3$\times$3 convolutional layer) and adaptively calculates depth bin centers based on predicted depth range parameters and normalized bin widths. Subsequently, final dense depth maps are generated via a linear combination of the estimated depth bin probabilities and the corresponding bin centers.

The decoder first utilizes $r$ and $w$ to calculate the physical center depth values $c = \{c_1, c_2, \dots, c_N\}$. Specifically, $w$ is converted into relative widths $b_i$ (summing to 1) via a softmax function:
\begin{equation}
    b_i = \frac{e^{w_i}}{\sum_{j=1}^{N} e^{w_j}}.
\end{equation}
Based on $r$, the scene depth range $[d_{\min}, d_{\max}]$ is determined to calculate the boundaries and centers of each bin:
\begin{equation}
    c_i = d_{\min} + (d_{\max} - d_{\min}) \left( \frac{b_i}{2} + \sum_{j=1}^{i-1} b_j\right).
\end{equation}
Simultaneously, the decoder adopts a bottom-up (low-to-high resolution) fusion strategy. During each upsampling stage, it not only upsamples features from deeper layers but also fuses feature maps from the corresponding encoder scale via skip connections. This ensures that spatial details from low-level features effectively guide the structural information of high-level features, progressively refining the depth prediction to output a probability volume $P \in \mathbb{R}^{H \times W \times N}$. Here, $P_k(p)$ denotes the probability that pixel $p$ belongs to the $k$-th depth bin. This classification-based formulation allows the network to model depth uncertainty and produce sharper object boundaries compared to direct regression. Finally, the continuous depth map $D_t$ is obtained by the soft-weighted combination of the adaptive bin centers from the DPV. The depth at pixel $(u,v)$ is calculated as:
\begin{equation}
    D_t(u,v) = \sum_{i=1}^{N} P_i(u,v) \cdot c_i.
\end{equation}

\subsection{Pose Network and Self-Supervised Training}\label{sec3.3}

Following standard practices in self-supervised depth estimation~\cite{ref11}, we use a ResNet-18 based architecture for the Pose Network. It accepts a pair of images $(I_t, I_{t'})$ concatenated along the channel dimension. The network predicts the 6-DoF relative camera pose $T_{t \to t'}$, consisting of a 3D translation vector and a 3D rotation vector (represented by axis-angle). This pose is used to project the source image pixels onto the target image plane.
Our model is trained in a self-supervised manner using a combination of photometric reprojection loss and edge-aware smoothness loss.

\textbf{Photometric Loss.} 
Based on the predicted depth $D_t$ and the estimated pose $T_{t \to t'}$, we synthesize the target image $I_{t' \to t}$ by warping the source image $I_{t'}$. The photometric loss $L_p$ measures the similarity between the original target image $I_t$ and the synthesized image $I_{t' \to t}$. As introduced in~\cite{ref11}, we use a combination of the L1 norm and the Structural Similarity Index (SSIM):
\begin{equation}
    L_p = \min_{t'} \left( \alpha \frac{1 - \text{SSIM}(I_t, I_{t' \to t})}{2} + (1 - \alpha) \| I_t - I_{t' \to t} \|_1 \right),
\end{equation}
where $\alpha$ is set to 0.85 empirically~\cite{ref11}. In addition, to deal with out-of-view pixels and occluded objects, we compute the minimum photometric loss across all source frames~\cite{ref11}.

\textbf{Smoothness Loss.} 
To encourage local smoothness of the predicted depth map while preserving object edges, we employ an edge-aware smoothness loss $L_s$, calculated as follows~\cite{ref11}:
\begin{equation}
    L_s = |\partial_x d_t^*| e^{-|\partial_x I_t|} + |\partial_y d_t^*| e^{-|\partial_y I_t|},
\end{equation}
where $d_t^* = D_t / \bar{D}_t$ denotes the mean-normalized inverse depth.

\textbf{Total Loss.} 
The final objective function is the weighted sum of the photometric loss and the smoothness loss:
\begin{equation}
    L_{total} = L_p + \lambda L_s.
\end{equation}
where $\lambda$ is a weighting factor (typically $10^{-3}$) to balance the two terms.

\section{Experiments}\label{sec4}

\subsection{Implementation Details}\label{sec4.1}

Our framework is implemented in PyTorch and trained on an NVIDIA RTX 4090 GPU for 20 epochs using AdamW. The learning rate uses a 3-epoch linear warmup to $1 \times 10^{-4}$, followed by a cosine decay to $1 \times 10^{-6}$. We set the number of adaptive depth bins to 256. The disparity smoothness weight is $10^{-3}$, and the photometric loss combines SSIM and L1 with weights of 0.85 and 0.15, respectively.

\subsection{Datasets and Evaluation Metrics}\label{sec4.2}
\textbf{KITTI.} The KITTI dataset \cite{ref1} is the standard benchmark for outdoor depth estimation. We adopt the data split of Eigen et al. \cite{ref42}, which contains 39,180 monocular triplets for training, 4,424 for validation, and 697 for testing. 

\textbf{Make3D.} To evaluate the generalization ability of our model, we use the Make3D dataset \cite{ref29} for zero-shot testing. It consists of 134 test images representing outdoor scenes.

\textbf{Metrics.} We report the results using the standard metrics used in prior works \cite{ref11}: absolute relative error (Abs Rel), squared relative error (Sq Rel), root mean squared error (RMSE), root mean squared logarithmic error (RMSE log), and threshold accuracy ($\delta < 1.25, 1.25^2, 1.25^3$).

\textbf{Baselines.} Due to variations in evaluation protocols across existing literature, we carefully select representative self-supervised approaches to benchmark our model's performance. These baselines span a wide spectrum of architectures, ranging from computationally intensive frameworks built upon Vision Transformers for dense prediction (ViT)~\cite{ref16} backbones to highly efficient lightweight networks. Furthermore, we extend our comparison to include recent state-of-the-art foundation models, such as Depth Anything~\cite{ref45}, to explicitly demonstrate the necessity and efficacy of in-domain self-supervised training.

\begin{table*}[htbp]
\centering
\caption{\textbf{Quantitative comparison on the KITTI Eigen split.} Comparison with other self-supervised monocular methods. Methods are grouped by training mode: MS indicates monocular and stereo training, and M indicates monocular training. The best results are highlighted in \textbf{bold}.}
\label{tab:kitti_comparison}
\setlength{\tabcolsep}{6pt}
\resizebox{\textwidth}{!}{%
    \begin{tabular}{l|c|cccc|ccc}
    \toprule
    \multicolumn{1}{c|}{Method} & Mode & RMSE$\downarrow$ & RMSE log$\downarrow$ & Sq Rel$\downarrow$ & Abs Rel$\downarrow$ & $\delta < 1.25\uparrow$ & $\delta < 1.25^2\uparrow$ & $\delta < 1.25^3\uparrow$ \\
    \midrule
    Monodepth2 \cite{ref11}& MS & 4.750 & 0.196 & 0.818 & 0.106 & 0.874 & 0.957 & 0.979 \\
    DepthHints \cite{ref33}& MS & 4.627 & 0.189 & 0.769 & 0.105 & 0.875 & 0.959 & 0.982 \\
    HR-Depth \cite{ref7}     & MS & 4.612 & 0.185 & 0.785 & 0.107 & 0.887 & 0.962 & 0.982 \\
    R-MSFM6 \cite{ref34}& MS & 4.625 & 0.189 & 0.787 & 0.111 & 0.882 & 0.961 & 0.981 \\
    \midrule
    Monodepth2 \cite{ref11}& M  & 4.863 & 0.193 & 0.903 & 0.115 & 0.877 & 0.959 & 0.971 \\
    HR-Depth \cite{ref7}     & M  & 4.632 & 0.185 & 0.792 & 0.109 & 0.884 & 0.962 & 0.983 \\
    CADepth-Net \cite{ref35}& M  & 4.535 & 0.181 & 0.769 & 0.105 & 0.892 & 0.964 & 0.983 \\
    DIFFNet \cite{ref36}& M  & 4.483 & 0.180 & 0.764 & 0.102 & 0.896& \textbf{0.965}& 0.983 \\
    MonoFormer \cite{ref37}& M  & 4.580 & 0.183 & 0.846 & 0.104 & 0.891 & 0.962 & 0.982 \\
    SRD-Depth \cite{ref38}& M  & 4.619 & 0.186 & 0.762 & 0.111 & 0.877 & 0.961 & 0.983 \\
    Lite-Mono \cite{ref12}& M  & 4.561 & 0.183 & 0.765 & 0.107 & 0.886 & 0.963 & 0.983 \\
    Liu \textit{et al.} \cite{ref15}& M & 4.724 & 0.187 & 0.747 & 0.114 & 0.863 & 0.960 & \textbf{0.984}\\
    Dynamo-Depth \cite{ref39}& M  & 4.505 & 0.183 & 0.758 & 0.112 & 0.873 & 0.959 & \textbf{0.984}\\
    PuriLight \cite{ref44}& M & 4.536 & 0.182 & 0.747 & 0.106 & 0.890 & 0.964 & \textbf{0.984}\\
    \textbf{SAP-Depth (Ours)}& M & \textbf{4.453}& \textbf{0.178} & \textbf{0.733}& \textbf{0.101} & \textbf{0.898}& \textbf{0.965}& 0.983 \\
    \bottomrule
    \end{tabular}%
}
\vspace{-0.4cm}
\end{table*}

\subsection{KITTI Results}\label{sec4.3}

\begin{figure*}[!t]
    \centering
    \includegraphics[width=0.92\linewidth]{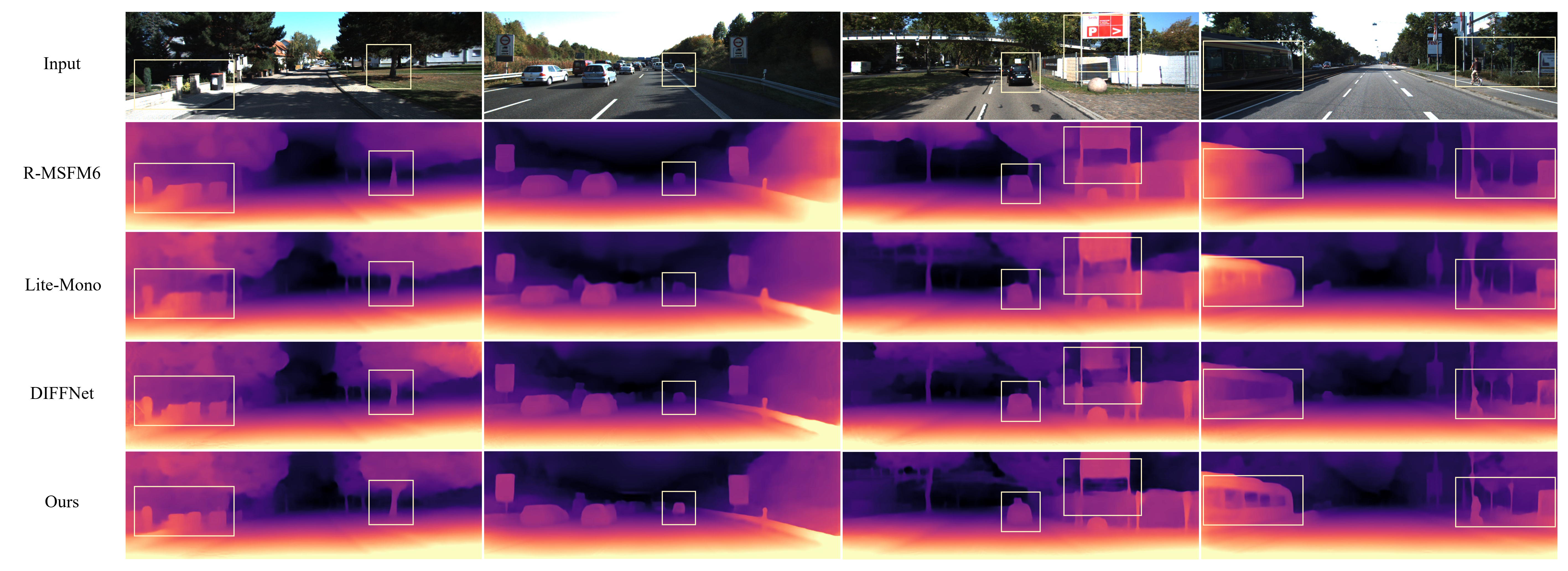}
    \caption{\textbf{Qualitative comparison on the KITTI dataset .} We compare our method against Lite-Mono\cite{ref12}, R-MSFM6\cite{ref34}, and DIFFNet\cite{ref36}. As shown, our model generates more complete depth predictions and preserves sharper object boundaries.}
    \label{fig:qualitative}
    \vspace{-0.4cm}
\end{figure*}

To intuitively demonstrate the superiority of our proposed method, we provide qualitative comparisons between SAP-Depth and other state-of-the-art methods on the KITTI dataset. Figure \ref{fig:qualitative} visualizes the depth maps generated by different models alongside the original RGB images.

As observed in the Figure \ref{fig:qualitative}, our method generates depth maps with sharper boundaries and more distinct structural details compared to baseline methods. Specifically, in regions with complex occlusion or thin structures (such as traffic signs and poles) , our model successfully preserves the geometric shapes, whereas other methods tend to produce blurred or over-smoothed artifacts. Furthermore,  our method demonstrates better consistency in depth inference for distant background objects. 

\subsection{Make3D Results}\label{sec4.4}

The proposed method is evaluated on the Make3D dataset to show its generalization ability in different outdoor scenes. The model trained on KITTI is directly inferred without any fine-tuning. Table \ref{tab:make3d_comparison} shows the comparison of SAP-Depth with the other methods, and SAP-Depth performs the best. 

\begin{table}[htbp]
\centering
\caption{\textbf{Zero-shot generalization results on the Make3D dataset.} Models are trained on KITTI and tested on Make3D without fine-tuning. The best results are highlighted in \textbf{bold}.}
\label{tab:make3d_comparison}
\setlength{\tabcolsep}{2pt}
\begin{tabular}{l|c|c|cccc}
\toprule
\multicolumn{1}{c|}{Method} & Size & Mode & RMSE$\downarrow$ & RMSE$_{log}\downarrow$ & Sq Rel$\downarrow$ & Abs Rel$\downarrow$ \\
\midrule
Monodepth2 \cite{ref11}& $640 \times 192$ & M & 7.418 & 0.163 & 3.589 & 0.322 \\
HR-Depth \cite{ref7}     & $640 \times 192$ & M & 7.024 & 0.159 & 3.208 & 0.315 \\
CADepth-Net \cite{ref35}& $640 \times 192$ & M & 7.066 & 0.159 & 3.086 & 0.312 \\
DIFFNet \cite{ref36}& $640 \times 192$ & M & 7.008 & 0.155 & 3.313 & 0.309 \\
Lite-Mono \cite{ref12}& $640 \times 192$ & M & 6.981 & 0.158 & 3.060 & 0.305 \\
Zhao \textit{et al.} \cite{ref40}& $640 \times 192$ & M & 7.095 & 0.158 & 3.200 & 0.316 \\
Xiong \textit{et al.} \cite{ref41}& $640 \times 192$ & M & 7.005 & 0.161 & 3.102 & 0.319 \\
\textbf{SAP-Depth (Ours)}& $640 \times 192$ & M & \textbf{6.709}& \textbf{0.151}& \textbf{2.806}& \textbf{0.291} \\
\bottomrule
\end{tabular}
\vspace{-0.4cm}
\end{table}

\subsection{Ablation Study on Model Architectures}\label{sec4.5}

To further verify the effectiveness of the proposed SAP-Depth, we conduct a comprehensive ablation study to analyze the importance of different architectural designs. Specifically, we investigate the contributions of the LP Predictor and the HAP-FAM. We remove or adjust these modules and report their performance on the KITTI dataset, as summarized in Table \ref{tab:ablation}.

\textbf{The benefit of BPD decoder.} To verify the necessity of our discretization design, the proposed binning strategy is substituted with a standard continuous regression head. As reported in Table \ref{tab:ablation}, this removal leads to a significant accuracy drop. This highlights the limitation of direct regression, which often suffers from the ``regression-to-mean'' problem and fails to capture high-frequency details. Conversely, the BPD decoder discretizes the depth space into probabilistic bins, mitigating optimization difficulties and yielding sharper geometric structures.

\textbf{The benefit of LP Predictor.} Removing the dynamic binning mechanism causes a substantial performance degradation, suggesting that standard static inference fails to capture intricate depth variations across diverse environments. In contrast, the LP Predictor acts as a learnable perceiver, actively aligning the depth probability distribution with the perceived scene context for better adaptability.

\textbf{The benefit of HAP-FAM.} Substituting HAP-FAM with basic upsampling and concatenation leads to inferior performance across all metrics (Table \ref{tab:ablation}). This underscores the limitation of simple linear aggregation in bridging the semantic gap. Conversely, HAP-FAM effectively captures complex cross-scale correlations, selectively fusing high-level semantics with low-level structural details for accurate depth inference.

It is worth noting that, as summarized in Table \ref{tab:model_efficiency}, we further benchmark SAP-Depth against recent foundation models, namely Depth Anything\cite{ref45} and MiDaS v3.1\cite{ref46}, under a zero-shot setting on the KITTI dataset. However, we must clarify that MiDaS does not strictly adhere to the zero-shot protocol, as its massive training corpus intrinsically encompasses KITTI images. Despite this unfair advantage, our SAP-Depth significantly outperforms MiDaS across all evaluated metrics. Furthermore, while Depth Anything achieves commendable performance improvements, its heavyweight ViT architecture imposes prohibitive demands on both training resources and edge-side computational capacities. In contrast, SAP-Depth maintains an optimal balance, delivering highly competitive accuracy with a fraction of the computational footprint, making it highly suitable for real-world edge deployments.

\begin{table*}[t!]
    \centering
    \begin{minipage}[t]{0.48\linewidth}
        \centering
        \caption{\textbf{Ablation study on KITTI.} We validate our components by reverting to baselines: direct regression (w/o BPD), static prediction (w/o LP), and simple upsampling (w/o HAP-FAM). Best results are \textbf{bold}.}
        \label{tab:ablation}
        \vspace{0.1cm} 
        \resizebox{\linewidth}{!}{%
        \begin{tabular}{lccc}
            \toprule
            \multirow{2}{*}{Architecture} & \multicolumn{2}{c}{Error metrics $\downarrow$} & Accuracy metrics $\uparrow$ \\
            \cmidrule(lr){2-3} \cmidrule(lr){4-4}
             & Abs Rel & RMSE & $\delta < 1.25$ \\
            \midrule
            w/o BPD decoder & 0.106 & 4.580 & 0.886 \\
            w/o LP Predictor & 0.102 & 4.489 & 0.896 \\
            w/o HAP-FAM & 0.102 & 4.492 & \textbf{0.898} \\
            \textbf{SAP-Depth full model} & \textbf{0.101} & \textbf{4.453} & \textbf{0.898} \\
            \bottomrule
        \end{tabular}%
        }
    \end{minipage}
    \hfill 
    \begin{minipage}[t]{0.48\linewidth}
        \centering
        \caption{\textbf{Comparison against Foundation Models.} We benchmark the model complexity (Params and FLOPs) and estimation accuracy of SAP-Depth against large-scale pre-trained depth foundation models.}
        \label{tab:model_efficiency}
        \vspace{0.1cm}
        \resizebox{\linewidth}{!}{%
        \begin{tabular}{lcccc}
            \toprule
            \multirow{2}{*}{Method} & \multicolumn{2}{c}{Model Size $\downarrow$} & \multicolumn{2}{c}{Performance} \\
            \cmidrule(lr){2-3} \cmidrule(lr){4-5}
             & Params (M) & FLOPs (G) & Abs Rel $\downarrow$ & $\delta < 1.25 \uparrow$ \\
            \midrule
            MiDaS v3.1  \cite{ref46} & 345 & 174 & 0.127 & 0.850 \\
            Depth Anything  \cite{ref45}& 97.5 & 75.0 & \textbf{0.080} & \textbf{0.939} \\
            \textbf{SAP-Depth (Ours)} & \textbf{10.16} & \textbf{15.92} & 0.101 & 0.898 \\
            \bottomrule
        \end{tabular}%
        }
    \end{minipage}
\end{table*}

\subsection{Ablation Study on Number of Bins}\label{sec4.6}

To rigorously evaluate the impact of the discretization resolution, we investigate how the number of dynamic bins ($N$) affects the depth estimation performance. As presented in Table \ref{tab:bins_quantity}, we conduct experiments ranging from 16 to 256 bins. 

Increasing the number of bins generally enhances the resolution of the depth probability distribution. When scaling from 16 to 64 bins, the metrics show a consistent improvement, with RMSE reaching a local minimum of 4.452 at $N=64$. This indicates that a finer discretization initially allows the network to capture more precise depth transitions. Interestingly, at 128 bins, the model experiences a slight performance fluctuation. This phenomenon suggests that naively increasing bin density can temporarily introduce optimization ambiguities, where the network struggles to confidently assign probabilities to overly narrow, adjacent bins without sufficient contextual constraints. 

However, when the resolution is further scaled to $N=256$, the model successfully overcomes this bottleneck and achieves the best overall performance. Notably, the 256-bin configuration significantly minimizes the Sq Rel metric to a remarkable 0.733 and achieves the highest threshold accuracy ($\delta < 1.25$ of 0.898). Since Sq Rel strongly penalizes large outlier errors, this substantial improvement indicates that 256 bins are exceptionally effective at resolving complex boundary ambiguities and distant structures. Although the 64-bin setting yields a marginally lower RMSE (4.452 vs. 4.453), the 256-bin model dominates in logarithmic depth errors (RMSE$_{log}$) and accuracy, demonstrating a much more robust and holistic structural preservation. 

Finally, formulating the output as a dense probability volume $P \in \mathbb{R}^{H \times W \times N}$ introduces linear growth in the spatial-channel footprint. Scaling beyond 256 bins inherently exceeds our current available computational resources and GPU memory constraints during training. Therefore, configuring $N=256$ yields the optimal balance, maximizing structural detail preservation while remaining computationally feasible.

\begin{table}[htbp]
\centering
\caption{Performance evaluation across different numbers of adaptive bins ($N$). Setting $N=256$ achieves the optimal accuracy under the available computational constraints.}\label{tab:bins_quantity}
\setlength{\tabcolsep}{4pt}
\begin{tabular}{lcccccc}
\toprule
Number of Bins ($N$) & RMSE $\downarrow$ & RMSE$_{log} \downarrow$ & Sq Rel $\downarrow$ & Abs Rel $\downarrow$ & $\delta < 1.25 \uparrow$ \\
\midrule
16 Bins  & 4.593 & 0.182 & 0.816 & 0.105 & 0.891 \\
32 Bins  & 4.524 & 0.180 & 0.773 & 0.103 & 0.895 \\
64 Bins  & \textbf{4.452} & 0.179 & 0.755 & \textbf{0.101} & 0.897\\
128 Bins & 4.534 & 0.181 & 0.769 & 0.102 & 0.896 \\
\textbf{256 Bins (Ours)} & 4.453 & \textbf{0.178} & \textbf{0.733} & \textbf{0.101} & \textbf{0.898} \\
\bottomrule
\end{tabular}
\vspace{-0.4cm}
\end{table}

\subsection{Model Efficiency and Complexity}\label{sec4.7}

 In Table \ref{tab:efficiency}, 
A critical factor for the real-world deployment of monocular depth estimation models is the delicate balance between computational complexity and prediction accuracy. In Table \ref{tab:efficiency}, we compare the parameter count (Params) and computational overhead (FLOPs) of our proposed SAP-Depth against representative state-of-the-art architectures.

It is worth noting that while extremely lightweight architectures like Lite-Mono exhibit lower parameter counts and computational costs through customized Lite-CNN backbones, this aggressive compression inherently trades off representational capacity, leading to sub-optimal structural accuracies (e.g., an Abs Rel of 0.107). Conversely, heavily parameterized models such as MonoFormer (159M parameters) and Monodepth2 utilizing a ResNet-50 backbone (32.50M parameters) incur substantial computational burdens. In contrast, our SAP-Depth strikes an optimal trade-off: it effectively prioritizes the rigorous preservation of high-resolution structural details by adopting the HRNet-18 encoder, achieving a superior Abs Rel of 0.101 while requiring merely 10.16M parameters and 15.92G FLOPs. Notably, SAP-Depth operates with fewer FLOPs than the ResNet-50 based Monodepth2 (16.70G) and requires less than half the computational cost of R-MSFM6 (34.5G).

As shown in the architectural analysis, despite incorporating the sophisticated dynamic bins mechanism, SAP-Depth maintains an elegant and highly efficient design through $1 \times 1$ feature compressions and global pooling operations. Out of the 10.16M total parameters, the shared HRNet-18 encoder accounts for the vast majority of the model capacity (approximately 9.57M parameters). In stark contrast, our customized Bin-wise Probability Distribution (BPD) decoder requires only around 0.45M parameters, and the LP Predictor introduces a negligible overhead of merely 0.14M parameters. This explicit parameter distribution evidently demonstrates that our method achieves significant performance gains through an intelligently formulated probabilistic framework rather than relying on a heavily parameterized, computationally expensive decoder.

\begin{table}[htbp]
\centering
\caption{Comparison of Model Size (Params) and Computational Complexity (FLOPs) evaluated on the $192 \times 640$ input resolution.}\label{tab:efficiency}
\setlength{\tabcolsep}{8pt}
\begin{tabular}{l|c|cc|c}
\toprule
Method & Backbone & Params (M) $\downarrow$ & FLOPs (G) $\downarrow$ & Abs Rel $\downarrow$ \\
\midrule
Monodepth2 \cite{ref11}& ResNet-50 & 32.50 & 16.70 & 0.110 \\
MonoFormer \cite{ref37} & ResNet-50 & 159 & 112 & 0.104 \\
Lite-Mono \cite{ref12} & Lite-CNN & \textbf{3.07} & \textbf{5.04} & 0.107 \\
R-MSFM6 \cite{ref34} & ResNet-18 & 5.24  & 34.5 & 0.109 \\
\midrule
\textbf{SAP-Depth (Ours)} & HRNet-18 & 10.16 & 15.92 & \textbf{0.101} \\
\bottomrule
\end{tabular}
\end{table}

\section{Conclusion}

In this paper, we proposed SAP-Depth, a novel self-supervised monocular depth estimation architecture designed to enhance object details and scene adaptation. By synergistically integrating dynamic Bins with a HAP-FAM within a BPD decoder, our method achieves adaptive depth interval fitting and refined cross-scale feature representation. Extensive experiments on the KITTI dataset and zero-shot testing on Make3D validate the superiority of our approach, demonstrating that SAP-Depth effectively captures fine-grained details and maintains stable accuracy in complex scenarios, thereby offering a reliable solution for real-world applications.

\subsubsection*{Disclosure of Interests}
The authors have no competing interests to declare that are relevant to the content of this article.


\begin{thebibliography}{99}
\setlength{\itemsep}{0pt}
\setlength{\parskip}{0pt}
\small

\bibitem{ref1}
Geiger, Andreas, et al. Vision meets robotics: The kitti dataset. \textit{The international journal of robotics research} 32.11 (2013): 1231-1237. 

\bibitem{ref2}
Newcombe, Richard A., et al. Kinectfusion: Real-time dense surface mapping and tracking. \textit{2011 10th IEEE international symposium on mixed and augmented reality}. Ieee, 2011. 

\bibitem{ref3}
Luo, Xuan, et al. Consistent video depth estimation. \textit{ACM Transactions on Graphics (ToG)} 39.4 (2020): 71-1. 

\bibitem{ref4}
Dudek, Gregory, and Michael Jenkin. \textit{Computational principles of mobile robotics}. Cambridge university press, 2024. 

\bibitem{ref5}
Godard, Clément, Oisin Mac Aodha, and Gabriel J. Brostow. Unsupervised monocular depth estimation with left-right consistency. \textit{Proceedings of the IEEE conference on computer vision and pattern recognition}. 2017. 

\bibitem{ref7}
Lyu, Xiaoyang, et al. Hr-depth: High resolution self-supervised monocular depth estimation. \textit{Proceedings of the AAAI conference on artificial intelligence}. Vol. 35. No. 3. 2021. 

\bibitem{ref8}
Li, Zhenyu, et al. Binsformer: Revisiting adaptive bins for monocular depth estimation. \textit{arXiv preprint arXiv:2204.00987} (2022). 

\bibitem{ref9}
Bhat, Shariq Farooq, Ibraheem Alhashim, and Peter Wonka. Localbins: Improving depth estimation by learning local distributions. \textit{European Conference on Computer Vision}. Cham: Springer Nature Switzerland, 2022. 

\bibitem{ref10}
Bhat, Shariq Farooq, et al. Zoedepth: Zero-shot transfer by combining relative and metric depth. \textit{arXiv preprint arXiv:2302.12288} (2023). 

\bibitem{ref11}
Godard, Clément, et al. Digging into self-supervised monocular depth estimation. \textit{Proceedings of the IEEE/CVF international conference on computer vision}. 2019. 

\bibitem{ref12}
Zhang, Ning, et al. Lite-mono: A lightweight cnn and transformer architecture for self-supervised monocular depth estimation. \textit{Proceedings of the IEEE/CVF conference on computer vision and pattern recognition}. 2023. 

\bibitem{ref13}
Garg, Ravi, et al. Unsupervised cnn for single view depth estimation: Geometry to the rescue. \textit{European conference on computer vision}. Cham: Springer International Publishing, 2016. 

\bibitem{ref14}
Zhu, Shengjie, Garrick Brazil, and Xiaoming Liu. The edge of depth: Explicit constraints between segmentation and depth. \textit{Proceedings of the IEEE/CVF conference on computer vision and pattern recognition}. 2020. 

\bibitem{ref15}
Runze Liu, et al. Unsupervised monocular depth estimation based on hierarchical feature-guided diffusion. arXiv preprint arXiv:2406.09782, 2024.  

\bibitem{ref16}
Ranftl, René, Alexey Bochkovskiy, and Vladlen Koltun. "Vision transformers for dense prediction." Proceedings of the IEEE/CVF international conference on computer vision. 2021.

\bibitem{ref17}
Godard C, Mac Aodha O, Brostow G J. Unsupervised monocular depth estimation with left-right consistency[C]//Proceedings of the IEEE conference on computer vision and pattern recognition. 2017: 270-279.

\bibitem{ref18}
Ke B, Obukhov A, Huang S, et al. Repurposing diffusion-based image generators for monocular depth estimation[C]//Proceedings of the IEEE/CVF conference on computer vision and pattern recognition. 2024: 9492-9502.

\bibitem{ref21}
Wang, Jiyuan, et al. Jasmine: Harnessing Diffusion Prior for Self-supervised Depth Estimation. \textit{arXiv preprint arXiv:2503.15905} (2025).  

\bibitem{ref22}
Bhat, Shariq Farooq, Ibraheem Alhashim, and Peter Wonka. Adabins: Depth estimation using adaptive bins. \textit{Proceedings of the IEEE/CVF conference on computer vision and pattern recognition}. 2021. 

\bibitem{ref29}
Saxena, Ashutosh, Min Sun, and Andrew Y. Ng. Make3d: Learning 3d scene structure from a single still image. \textit{IEEE transactions on pattern analysis and intelligence} 31.5 (2008): 824-840. 

\bibitem{ref30}
Guizilini, Vitor, et al. 3d packing for self-supervised monocular depth estimation. \textit{Proceedings of the IEEE/CVF conference on computer vision and pattern recognition}. 2020. 

\bibitem{ref33}
Watson, Jamie, et al. Self-supervised monocular depth hints. \textit{Proceedings of the IEEE/CVF international conference on computer vision}. 2019. 

\bibitem{ref34}
Zhou, Zhongkai, et al. R-msfm: Recurrent multi-scale feature modulation for monocular depth estimating. \textit{Proceedings of the IEEE/CVF international conference on computer vision}. 2021. 

\bibitem{ref35}
Yan, Jiaxing, et al. Channel-wise attention-based network for self-supervised monocular depth estimation. \textit{2021 International Conference on 3D vision (3DV)}. IEEE, 2021. 

\bibitem{ref36}
Zhou, Hang, David Greenwood, and Sarah Taylor. Self-supervised monocular depth estimation with internal feature fusion. arXiv preprint arXiv:2110.09482 (2021). 

\bibitem{ref37}
Zhao, Chuyang, et al. Monoformer: One transformer for both diffusion and autoregression. arXiv preprint arXiv:2409.16280 (2024). 

\bibitem{ref38}
Liu, Zhong, et al. Self-supervised monocular depth estimation with self-reference distillation and disparity offset refinement. \textit{IEEE transactions on circuits and systems for video technology} 33.12 (2023): 7565-7577. 

\bibitem{ref39}
Sun, Yihong, and Bharath Hariharan. Dynamo-depth: Fixing unsupervised depth estimation for dynamical scenes. \textit{Advances in Neural Information Processing Systems} 36 (2023): 54987-55005. 

\bibitem{ref40}
Zhao, Hailiang, et al. Learning effective geometry representation from videos for self-supervised monocular depth estimation. \textit{ISPRS International Journal of Geo-Information} 13.6 (2024): 193. 

\bibitem{ref41}
Xiong, Mingkang, et al. Monocular depth estimation using self-supervised learning with more effective geometric constraints. \textit{Engineering Applications of Artificial Intelligence} 128 (2024): 107489. 

\bibitem{ref42}
Eigen, David, Christian Puhrsch, and Rob Fergus. Depth map prediction from a single image using a multi-scale deep network. \textit{Advances in neural information processing systems} 27 (2014). 

\bibitem{ref43}
Fu, Huan, et al. Deep ordinal regression network for monocular depth estimation. \textit{Proceedings of the IEEE conference on computer vision and pattern recognition}. 2018.

\bibitem{ref44}
Chen, Yujie, et al. PuriLight: A Lightweight Shuffle and Purification Framework for Monocular Depth Estimation. \textit{arXiv preprint arXiv:2602.11066} (2026).

\bibitem{ref45}
Yang, Lihe, et al. "Depth anything: Unleashing the power of large-scale unlabeled data." Proceedings of the IEEE/CVF conference on computer vision and pattern recognition. 2024.

\bibitem{ref46}
Birkl, Reiner, Diana Wofk, and Matthias Müller. "Midas v3. 1--a model zoo for robust monocular relative depth estimation." arXiv preprint arXiv:2307.14460 (2023).

\end{thebibliography}
\end{document}